\documentclass[runningheads]{llncs}
\usepackage{graphicx}
\usepackage{amsmath}
\usepackage{mathbbol}
\usepackage{orcidlink} 
\usepackage{color}

\usepackage[normalem]{ulem}

\begin{document}

\title{Muscle volume quantification: guiding transformers with anatomical priors\thanks{\scriptsize{This work has been supported by the European Regional Development Fund (FEDER), the Pays de la Loire region on the Connect Talent scheme (MILCOM Project), Nantes Métropole (Convention 2017-10470) and FULGUR (Team: 30 researchers, Grant:1.9 M€)  which benefits from a French Research Agency aid (reference ANR-19-STPH-0003). FULGUR is part of the perspective of the Paris 2024 Olympic and Paralympic Games in collaboration with the French Federations of Athletics, Rugby and Ice Sports, Universities of Nantes, Côte d’Azur, Savoie Mont Blanc, Jean Monnet Saint-Etienne, Saclay, the Mines Saint Etienne school, the french Alternative Energies and Atomic Energy Commission (CEA), the French National Centre for Scientific Research (CNRS), Natural Grass and Super Sonic Imagine.}}}

\author{Louise Piecuch\inst{1,2} \orcidID{0009-0004-0192-1220} \and
Vanessa Gonzales Duque\inst{1,3} \orcidID{0000-0002-2149-6581} \and
Aurélie Sarcher \inst{2} \orcidID{0000-0002-6408-6291} \and
Enzo Hollville \inst{5,6} \orcidID{0000-0001-6277-8051} \and
Antoine Nordez \inst{2,4} \orcidID{0000-0002-7276-4793}\and
Giuseppe Rabita \inst{5} \orcidID{0000-0002-0548-3019}\and
Gaël Guilhem \inst{5}\orcidID{0000-0002-0377-2060}\and
Diana Mateus\inst{1} \orcidID{0000-0002-2252-8717}}
%index{Piecuch Louise} %index{Gonzales Duque Vanessa} %index{Sarcher Aurélie} %index{Hollville Enzo} %index{Nordez Antoine} %index{Rabita Giuseppe} %index{Guilhem Gaël} %index{Mateus Diana} 
\authorrunning{Piecuch L. et al.}

\institute{Nantes Université, École Centrale Nantes LS2N, UMR CNRS 6004,   France
\email{Louise.Piecuch@ls2n.fr} \and
Nantes Université, Movement - Interactions - Performance, MIP, IP UR 4334 UFR STAPS,  Nantes - France \and
Technical University of Munich (TUM), Germany  \and
Institut Universitaire de France  (IUF),   Paris -  France \and
Laboratory Sport Expertise and Performance, INSEP, Paris -  France \and
Fédération Française de Badminton, Saint-Ouen - France
}
\maketitle   

\begin{abstract}

Muscle volume is a useful quantitative biomarker in sports, but also for the follow-up of degenerative musculo-skelletal diseases. In addition to volume, other shape biomarkers can be extracted by segmenting the muscles of interest from medical images. Manual segmentation is still today the gold standard for such measurements despite being very time-consuming. We propose a method for automatic segmentation of 18 muscles of the lower limb on 3D Magnetic Resonance Images to assist such morphometric analysis. By their nature, the tissue of different muscles is undistinguishable when observed in MR Images.
Thus, muscle segmentation algorithms cannot rely on appearance but only on contour cues. However, such contours are hard to detect and their thickness varies across subjects. To cope with the above challenges, we propose a segmentation approach based on a hybrid architecture, combining convolutional and visual transformer blocks. We investigate for the first time the behaviour of such hybrid architectures in the context of muscle segmentation for shape analysis. Considering the consistent anatomical muscle configuration, we rely on transformer blocks to capture the long-range relations between the muscles.
To further exploit the anatomical priors, a second contribution of this work consists in adding a regularisation loss based on an adjacency matrix of plausible muscle neighbourhoods estimated from the training data. 
Our experimental results on a unique database of elite athletes show it is possible to train complex hybrid models from a relatively small database of large volumes, while the anatomical prior regularisation favours better predictions.

\end{abstract}

\section{Introduction}

Skeletal muscles are composed of muscle fibers, usually arranged in bundles surrounded by connective tissue.
Different to other organs in the body, their shape can change relatively fast under physical training, injuries or under the effect of certain diseases. Therefore, the evolution, shape and volume of muscles have been studied in the sports and medical literature as biomarkers
\cite{Handsfield_et_al,Miller_et_al,Yokota_et_al,Sutherland_et_al}.
Such measurements can be extracted in a non-intrusive way through medical imaging.
Magnetic Resonance Imaging (MRI) is well suited for the task for its ability to image soft tissues with high contrast
\cite{Li_et_al,Miller_et_al}. 
An important intermediate step to go from images to biomarkers is the  segmentation. Once segmented it is possible to make comparisons between athletes, find trends according to discipline, sex, height, weight, or even within an individual, e.g. by detecting muscular asymmetries between the legs \cite{Handsfield_et_al,Sutherland_et_al}. In this study, our main focus is the sports domain.
However, muscle segmentation is also useful in the context of 
skeleto-muscular diseases like Duchenne's dystrophy, where monitoring muscle development is crucial. Therein, muscle segmentation helps track the disease progression, assess its impact on muscle tissue, or adapting treatment strategies. 

\begin{figure}[b]
\centering
\includegraphics[width=0.8\textwidth]{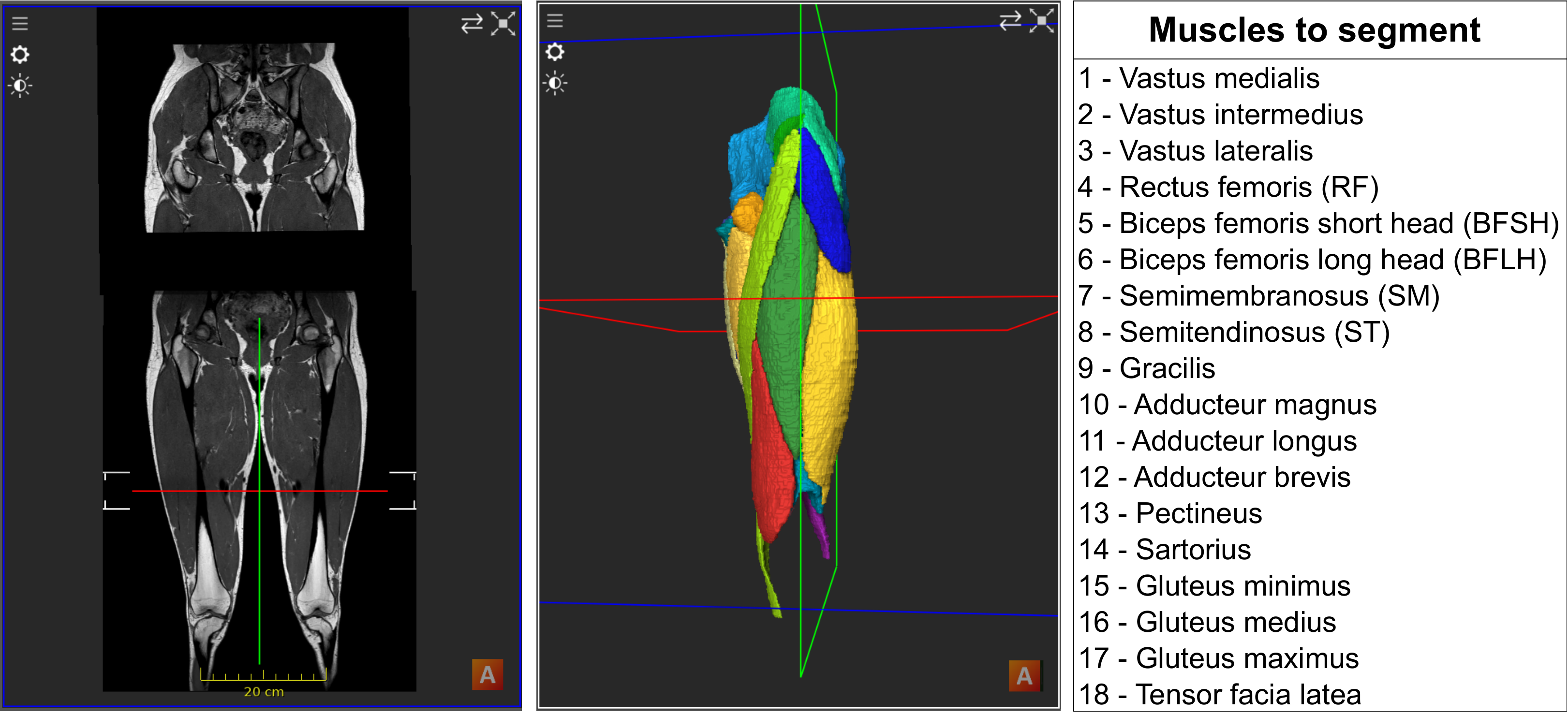}
\caption{Example of input MRI (\textbf{left}), and output labelmap (\textbf{middle}) The table (\textbf{right}) lists all the muscles to segment and their associated label. } \label{general_purpose}
\end{figure}

We aim to segment muscles from 3D MR images for subsequent shape analysis in elite sports (see Fig.~\ref{general_purpose}). 
Previous studies relied on manual annotations, which are very time-consuming and laborious \cite{Miller_et_al,Sutherland_et_al}. 
Indeed, segmenting MR images from elite athletes poses several challenges. In a broader context, the high number of muscles to be segmented is a major constraint. Also, these muscles are interconnected and influence each other. Unlike multi-organ segmentation where diverse labels are present, muscles have similar tissue types, so texture information is not discriminating.
Instead, muscle segmentation primarily relies on often thin or imperceptible boundaries. Elite athletes, with more developed muscles and less in-between fat, present an additional challenge. Finally, working with the full 3D data preserves important contextual information but necessitates memory management for large volumes.
These specific considerations highlight the need for tailored segmentation approaches that account for the distinct characteristics and requirements of sports-related imaging analysis. 
In order to automate the segmentation task while tackling these challenges, we rely on the UNETR architecture \cite{unetr} to leverage the strengths of both Convolutional Neural Netorks (CNN)s and transformers. Furthermore, medical and sports professionals have extensive knowledge of human anatomy, including muscle structure and their spatial collocation. Leveraging this expertise, we incorporate prior knowledge into the learning process through a regularization loss inspired by \cite{NonAdjloss} that enforces feasible muscle adjacencies. This loss leverages our knowledge of muscle anatomy to improve the accuracy and reliability of the segmentation model.

\section{Related Work}

Image segmentation is relevant in various sports-related applications. Miller et al. \cite{Miller_et_al}, examine and compare the variations in muscle volume between male elite sprinters and sub-elite sprinters. Furthermore, the study investigates the relationship between muscle volumes, strength, and sprint performances, all based on manual segmentations. The delineation of muscle boundaries and regions of interest are also performed by human experts in \cite{Sutherland_et_al} to characterize the hamstring muscles with a statistical shape model.
Alternative methods have emerged, including semi-automatic approaches. For instance, Hansdfield et al. \cite{Handsfield_et_al}, manually revised the output of an automatic algorithm to investigate the distribution of muscle volumes in the lower limb among elite sprinters. 
Gilles et al.\cite{Gilles_et_al}, focus on the registration and segmentation of hip and thigh muscles using deformable discrete models. More recently, automatic methods such as Yokota et al. \cite{Yokota_et_al} utilize multi-atlas techniques to automate the segmentation of hip and thigh muscles from CT scans. Cheng et al.~\cite{Cheng_et_al} rely on a U-Net for segmenting the quadriceps and patella from MRI scans, but primarily focusing on pediatric medical applications. 
Ni et al. \cite{Ni_et_al} proposed an automatic method for segmenting lower limb muscles of collegiate athletes (basketball, football, and soccer) using a cascaded 3-D CNN. The approach comprises two independently trained networks to address the muscle localization task on low-resolution images and the subsequent segmentation task on cropped high-resolution images.
To the best of our knowledge, apart from the methods mentioned earlier, there are only a limited number of automated approaches specifically designed for muscle segmentation. This is particularly true in the sports domain, where the task presents its own challenges.
However, given the amout of time required for a manual segmentation ($\sim$40h/subject in our case) there is a clear need to automate this task. Therefore, we propose an automatic method on a unique database of elite athletes, which is difficult to acquire and collect given the athletes' profiles, but also to annotate due to the significant muscle development and little adipose tissue.\\

The U-Net \cite{unet} architecture is considered the reference for automating segmentation tasks in medical imaging ~\cite{Li_et_al}.
However, the emergence of transformers in recent years has opened up new possibilities. While CNNs excel at capturing local structures, they have limitations when it comes to capturing long-range relationships among different regions in an image. As CNNs go deeper, their receptive field gradually expands, leading to distinct features extracted at different stages.
In contrast, transformer blocks leverage the power of Multi-head Self-attention (MSA) to establish a global receptive field, even at the lowest layer of models like the Vision Transformer (ViT). 
In this sense, transformer-based models, are well-suited for medical image segmentation since long-range dependencies are common within the human body. Another asset is the flexibility of their network architecture. Indeed, several architectures that combine transformers and CNNs have been proposed \cite{Li_et_al}, by offering various ways to integrate transformers into U-Net like networks. 
Petit et al. \cite{Petit_et_al} introduced transformers in the decoder of U-Net. Transformers can also be incorporated into the bottleneck section of the U-Net architecture, as demonstrated in TransUnet \cite{TransUnet}. Another approach involves independently processing the image through transformer blocks and convolutional layers, and subsequently merging the information obtained at each encoding step, as in \cite{CATS}. 
UNETR \cite{unetr} combines the strengths of CNNs and transformers by replacing the encoder of the U-Net architecture with a series of transformer blocks. A transformer block at the input reformulates the segmentation problem as a 1D sequence-to-sequence inference task, similar to transformers in natural language processing \cite{Li_et_al}. 
While more recent architectures have been deemed powerful, e.g.  Swin-Unetr \cite{swin_unter}, they are usually associated to 
higher training complexities, requiring larger computational resources and longer training times compared to traditional Transformers.
Based on our data limitations and Hasany et al.  \cite{Hasany_et_al} findings showing that UNETR captures global information fast, i.e., even at the third layer of transformers, we opt for a UNETR model.

As mentioned above, one of the challenges we face is that the tissue of different muscles appears identical when observed in MR Images, making texture information irrelevant.
On the contrary, contextual information can be a major asset. Multiple approaches can be employed to incorporate such relevant context, including modeling it with a loss function. 
Such functions can be constructed based on the morphology of the objects being segmented e.g. star-shaped~\cite{Mirikharaji_et_al} or vertebrae like~\cite{Arif_et_al}.
However, such shape priors are difficult to apply to our segmentation problem, since simple priors do not adequately capture the muscle variations and given there is no known atlas available. 
From a more topological perspective, BenTaieb et al. \cite{BenTaieb_et_al} addressed the issue of region exclusion and inclusion by penalizing incorrect label hierarchy. They noticed constraints between certain regions in their specific application and developed solutions to enforce inter-region connectivity. 
Since the muscles are separate entities, BenTaieb's method does not apply either.
Finally, Ganaye et al. \cite{NonAdjloss}, proposed a method based on multiple-organ and brain region adjacencies.  
Given that the positions of muscles remain consistent within the legs, the adjacency relationships among athletes' muscles are also expected to be preserved. Therefore, we adapt the idea of an adjacency constraint from \cite{NonAdjloss} to regularize the training of our automatic segmentation method.

\section{Methodology}
The purpose of this work is to design an automatic tool to segment muscles from MR images of the lower-limbs. More specifically, the approach receives as input 2 MR scans of the same subject (hip and thighs) and provides as output the semantic segmentation labelmaps (a probability of each voxel to belong to one of the 18 considered muscles), as shown in Fig.~\ref{general_purpose}. 
To address the above problem we rely on a hybrid (ViT + CNN) UNETR architecture \cite{unetr}, and in this way capture long range dependencies within a muscle and between muscles. To further reinforce the anatomical priors we first built an adjacency matrix from our training data, by estimating the probability of two muscles being next to each other. Then, we rely on this adjacency matrix to define a penalizing loss that forces the model to make predictions that respect the prior connectivities.
Next we describe the details of the architecture and loss.

\subsection{Model}\label{Model}
\begin{figure}[t!]
\centering
\includegraphics[width=0.7\textwidth]{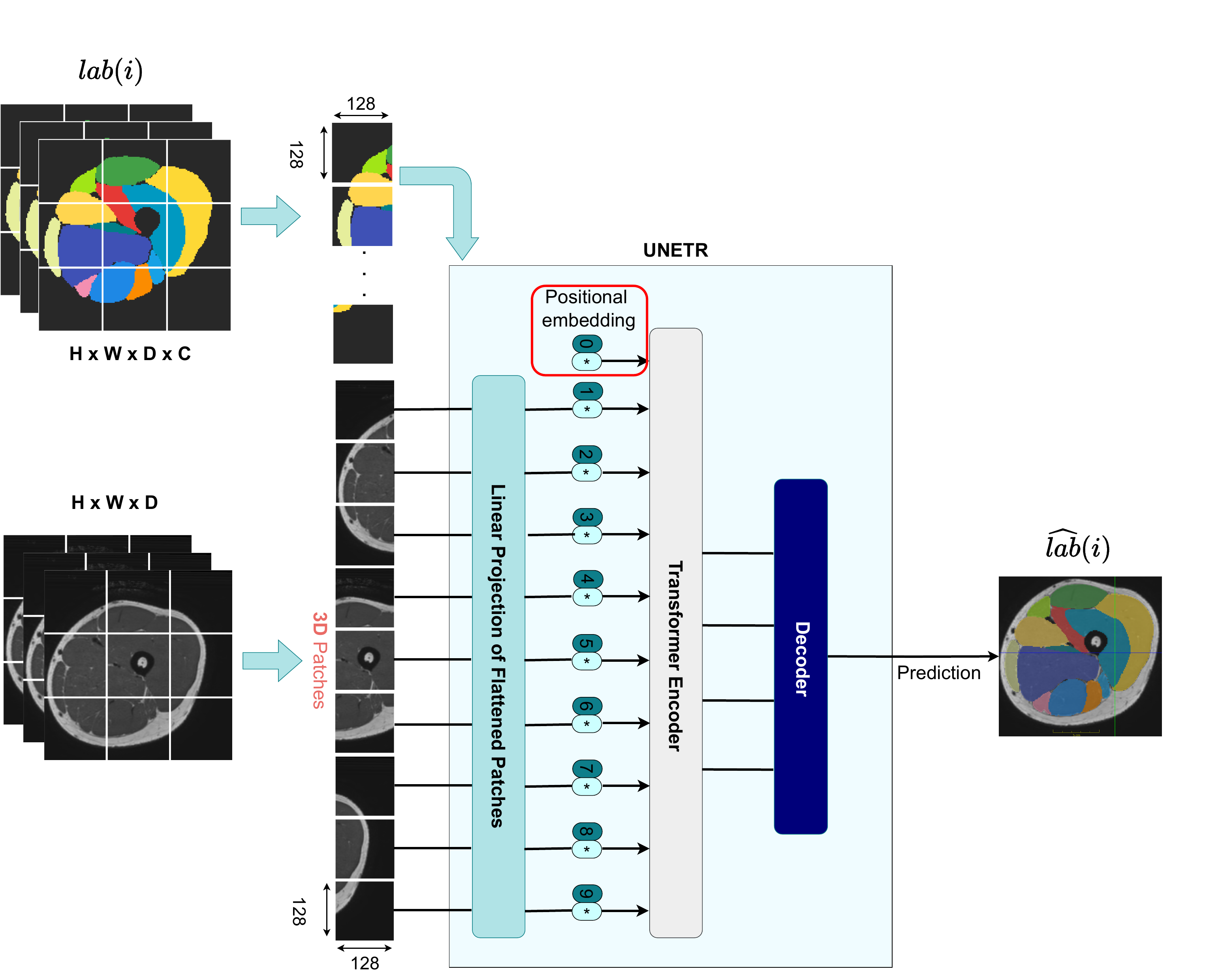}
\caption{General view of the method based on \cite{unetr} on muscle MRI.} \label{gen_architecture}
\end{figure}

Lets define the input to the model to be an image $\mathbf{x} \in \mathbb{R}^{H \times W \times D }$ (with $H \times W \times D$ the image size) and the associated ground truth labelmap as the function $\mathbf{lab}: i \in \mathbf{R}^{H \times W \times D} \longmapsto [0, ... , C] $, with $C$ the number of labels. We also denote as $ \widehat{lab}(i)$ the predicted labelmap obtained as output of the model. 
The chosen UNETR \cite{unetr} architecture is based on a U-Net, whose encoder has been replaced by a succession of $T$ transformer blocks. These $T$ blocks retain global information (e.g. on fairly long muscles), thanks to self-attention modules, while the architecture keeps
access to more local information through the convolutional layers. 
Next, we follow \cite{unetr} to describe details of each block.
Since transformer blocks work on 1D sequences, we convert our 3D input data 
$\mathbf{x}$ into a sequence of flattened non-overlapping patches $x^N_v$ of equal resolution ($P \times P \times P$), as shown in Fig ~\ref{gen_architecture}; thus, the sequence has length $N=(H \times W \times D) / P^3$. A linear layer, 
$E \in \mathbb{R}^{(P^3 . C) \times K }$, is then used to project each patch into a $K$ dimensional embedding space, which is the same throughout the transformers layers. A 1D learnable positional embedding $E_{pos} \in \mathbb{R}^{N \times K }$ is added to the sequence  of the projected patch embeddings in order to keep the spatial information and help reconstruct back the image. We denote the result of patch projections and positional embedding as:

\begin{equation}
    z_0=[x^1_v E;x^2_v E; ... ; x^N_v E]+E_{pos}
\end{equation}

\begin{figure}[t!]
\includegraphics[width=\textwidth]{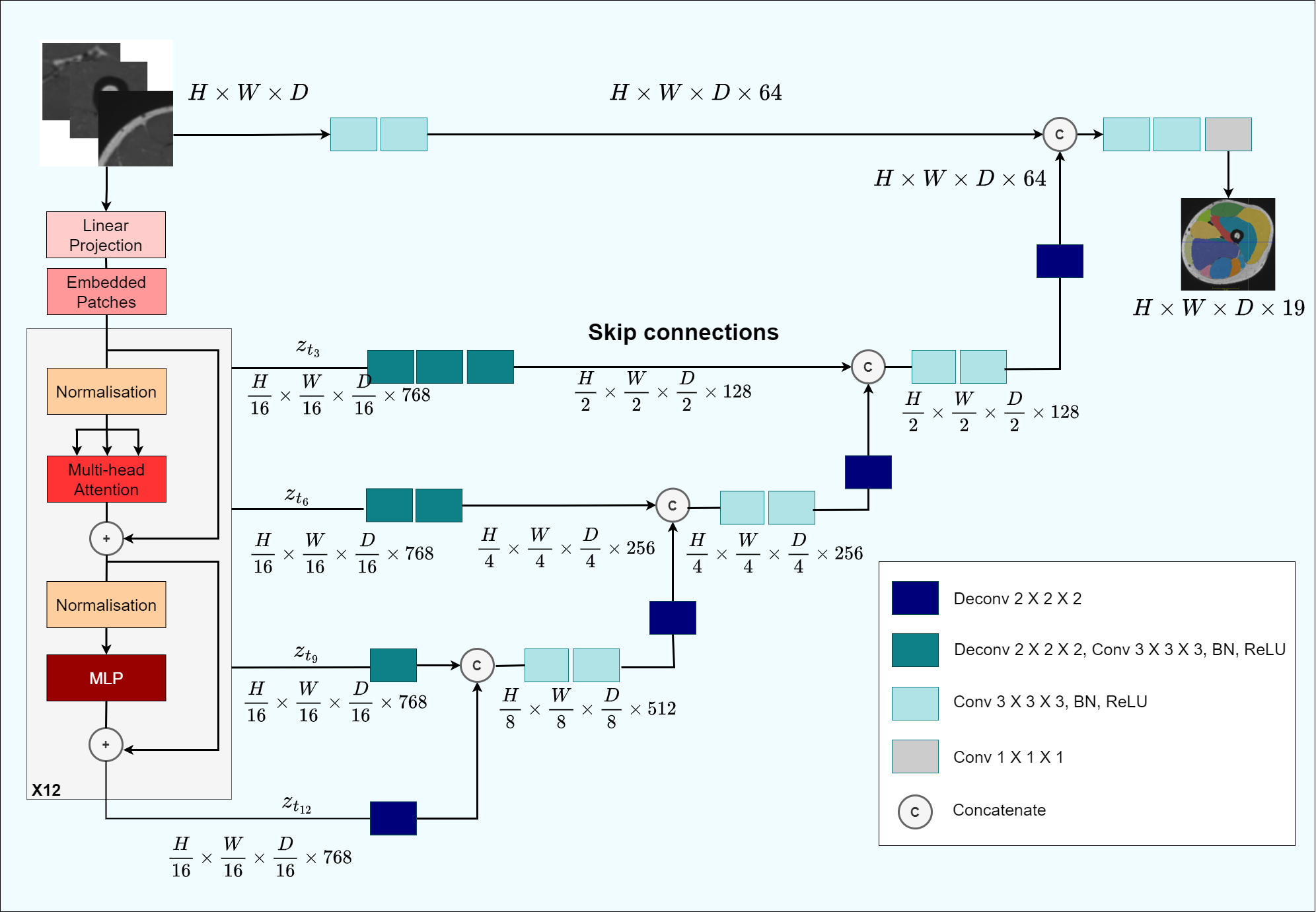}
\caption{Architecture of UNETR, modified figure from \cite{unetr}. } \label{architecture}
\end{figure}

After the embedding, the sequence $z_0$ passes through a stack of transformer blocks as shown in Fig~\ref{architecture}. A typical transformer block is composed of a multi-head self-attention (MSA) and a multi-layer perceptron (MLP) (c.f. Eq (6) in \cite{unetr}). The data is then passed through a normalisation layer, $Norm()$. 

\begin{equation}
    z_t'=MSA(Norm(z_{t-1})) + z_{t-1},  
    \qquad
    t=1...T,
\end{equation}
\begin{equation}
    z_t=MLP(Norm(z_{t}')) + z_{t}', 
    \qquad
    t=1...T,
\end{equation}

The UNETR architecture incorporates a direct connection between the transformer encoder and the CNN decoder block through skip-connections at different resolutions, enabling the calculation of the final semantic segmentation output. In the architecture bottleneck, a deconvolutional layer is employed to increase the resolution of the transformed feature map by a factor of 2. This upscaled feature map is then concatenated with the feature map from the previous transformer output (e.g. $z_{t_9}$ in Fig. ~\ref{architecture}).  
Next, consecutive $3 \times 3 \times 3$ convolutional layers are applied, followed by  an upsampling using a deconvolutional layer, until the output reaches the original input resolution. Finally, a $1 \times 1 \times 1$ convolutional layer with a softmax activation generates the voxel-wise semantic prediction.

\subsection{Prior anatomical knowledge}

The relative location of a muscle with respect to others is overall consistent across participants, especially for healthy subjects. Therefore, we propose to exploit such anatomical knowledge to further guide the training. In practice, we represent the relative positioning of the muscles with a probabilistic adjacency matrix. Similarly to \cite{NonAdjloss}, we employ this matrix within a regularizing loss term that penalizes predictions that do not respect the known adjacencies. \\

\begin{figure}[t!]
\centering
\includegraphics[width=0.7\textwidth]{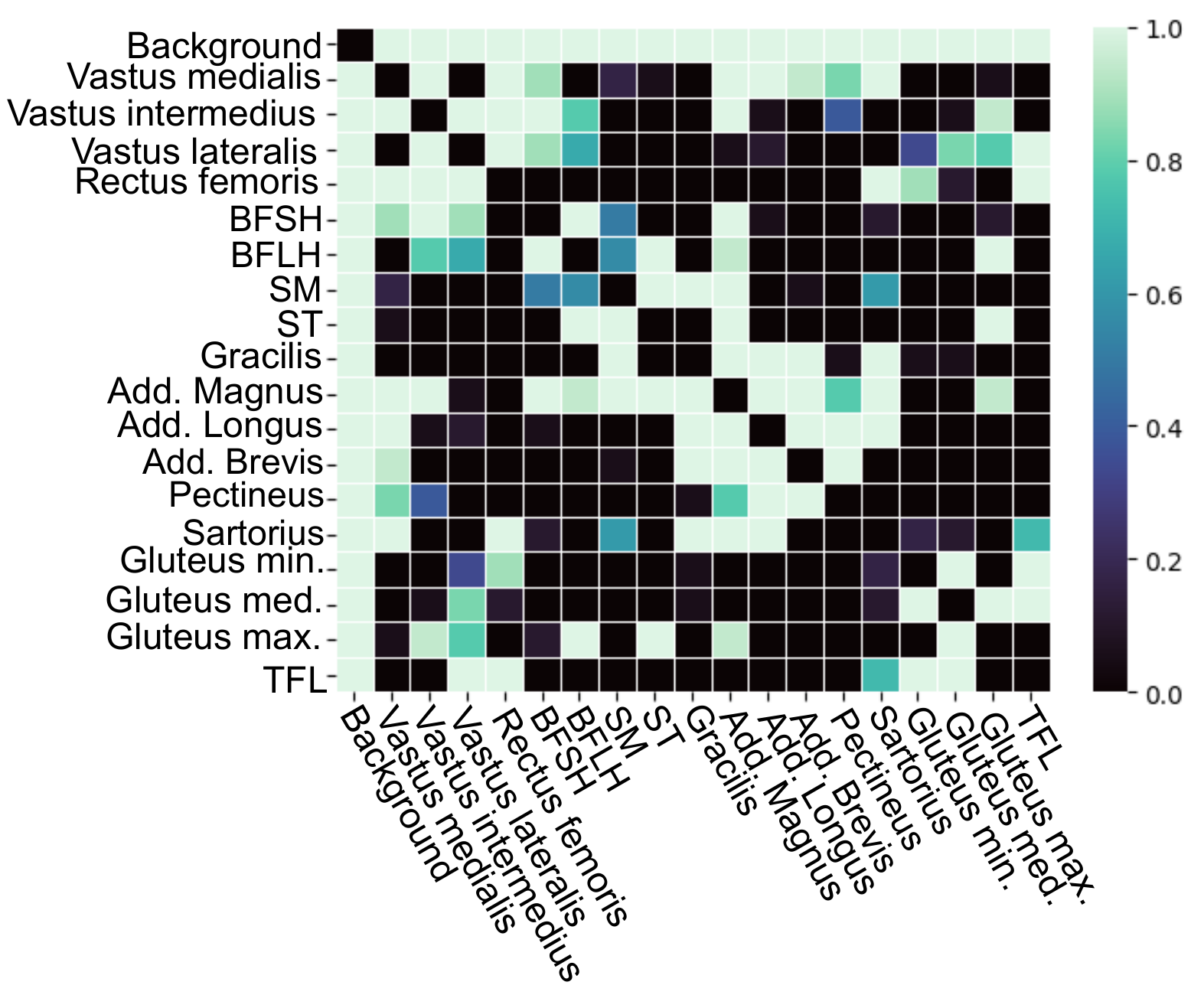}
\caption{Probabilistic adjacency matrix providing prior knowledge on muscle anatomy. The rows and columns of the matrix correspond to each label (muscles and background). Each element of the matrix has a value between 0 and 1. 
The higher the value is, the more likely the adjacency between 2 muscles.
}
\label{adj_matric}
\end{figure}

To create a probabilistic matrix, we extracted binary adjacency matrices for each subject in the database. To do this, three 4-neighbour filters are applied to the subject's manual segmentation labelmap $\mathbf{lab}$. 
These derivative filters perform the difference $d_{i,j}$ between the value $\mathbf{lab}(i)$ of a given voxel $i$ and its neighbour $j \in \mathcal{N}(i)$ (where $\mathcal{N}(i)$ corresponds to the neighborhood voxels of $i$), such that $d_{i,j}= \mathbf{lab}(i) - \mathbf{lab}(j)$. These filters are applied separately in the 3 directions of the labelmap. Any non-zero difference ($d_{i,j}\neq$0), indicates the presence of a boundary between these two neighboring voxels. Once $d_{i,j}$ has been calculated, and the boundaries found, we associate each boundary with the respective pair of labels (muscles). We then fill $1$ in the corresponding location of the $N_{muscles} \times N_{muscles}$ adjacency matrix if a boundary between a pair of labels was detected in any direction. After extracting the binary adjacency matrices for all subjects, we sum them up and normalise the result by the number of muscles. 
The resultant probabilistic adjacency matrix $ \mathbf{A}$ is shown in Fig.~\ref{adj_matric}. The process is summed up in the following equation: 

\begin{equation}\label{proba_matrix}
    a_{bc}(\mathbf{lab})=\sum_{i}\sum_{j \in \mathcal{N}(i)} \delta_{b,\mathbf{lab}(i)}\delta_{c,\mathbf{lab}(j)} ,
\end{equation}
where $b,c \in [0,...,C]$ are 2 different labels and $\delta_{b,\mathbf{lab}(i)}$ is the Kronecker delta function equal to $1$ when voxel $i$ has label $b$ (or $0$ otherwise).
Here, $a_{bc}$ is the adjacency function calculated on the ground-truth labelmap $\mathbf{lab}$ and $\tilde{a}_{bc}=(a_{bc}>0)$ is its binarized version, which is summed-up and normalised to obtain $\mathbf{A}$. The same function is computed during training but instead on the predicted probabilistic labelmap $\widehat{\mathbf{lab}}$, as discussed in Sec.~\ref{loss_func_sect}. \\

Unlike \cite{NonAdjloss}, we have chosen to keep the ground truth matrix probabilistic to consider the variability across subjects and the likelihood of the muscles connections.
Two muscles are considered adjacent if at least one of their voxels is in contact in the labelmap.

\subsection{Loss Function} \label{loss_func_sect}
The loss function combines a common segmentation loss function, the softdice cross-entropy loss with a regularization loss to consider the prior anatomical information of muscle adjacency.  

\begin{equation}
    L_{final}= L_{seg}(\mathbf{lab},\widehat{\mathbf{lab}}) + \lambda L_{NonAdjLoss}(\mathbf{lab},\widehat{\mathbf{lab}})
\end{equation}

A weighting lambda is applied to the regularization loss so as not to penalise the model too much while incorporating the anatomical constraint.

\paragraph{Soft Dice Cross Entropy Loss} is a combination of soft dice loss and cross-entropy:

\begin{equation}
 \resizebox{\textwidth}{!}
     {%
       $L_{seg}(\mathbf{lab},\widehat{\mathbf{lab}})=(1-\frac{2}{C}\sum_{c=1} ^{C} \frac{\sum_{i=1} ^{I} \mathbf{lab},\widehat{\mathbf{lab}}}{\sum_{i=1} ^{I} \mathbf{lab}^2 + \sum_{i=1} ^{I} \widehat{\mathbf{lab}}^2})(-\frac{1}{I} \sum_{i=1} ^{I} \sum_{c=1} ^{C} \mathbf{lab} \log \widehat{\mathbf{lab}}),$% 
    }
\end{equation}
where $I$ is the number of voxels ($I=H\times W\times D)$; $C$ is the number of classes; $\mathbf{lab}$ and $\widehat{\mathbf{lab}}$ denote respectively, the probabilistic prediction and ground-truth encoded in one-hot. 

\paragraph{NonAdjLoss} is the proposed regularisation loss enforcing the segmentation predictions to satisfy the anatomical constraints. 

\begin{equation}
    L_{NonAdjLoss}(\mathbf{lab},\widehat{\mathbf{lab}})= \sum_{\forall(b,c) \in [0, ..., C]}
    (1-a_{bc}(\mathbf{lab}))
    a_{bc}(\widehat{\mathbf{lab}}),
\end{equation}
where $a_{bc}(\widehat{\mathbf{lab}})$ is the adjacency function calculated during training from applying Eq. \ref{proba_matrix} to $\widehat{\mathbf{lab}}$. If the model predicts a wrong adjacency, we penalize it with the inverse of the probability of this link existing.
Thus, using the network with $a_{bc}(\widehat{\mathbf{lab}})$ as a differentiable adjacency matrix, allows us to penalize the forbidden connectivities of any prediction.

\section{Experimental Validation}
\subsection{Experimental Settings}

\textbf{Dataset.} The dataset, composed of 18 3D registered MRI (pelvis and thighs) of low-limb muscles from elite-athletes, was acquired at the  medical imaging centre of the INSEP. We split the dataset into 15:1:2 for training, validation and test.
The MR images were manually annotated to obtain the labelmaps of the 18 muscles in Fig. ~\ref{general_purpose}, which took between 30 to 40 hours per subject. The MRI are cropped to show only one leg for the training and inference. The average volume is $467.2\times 450.2 \times 1556.2$ pixels for a spacing of $0.55 \times 0.55 \times 0.55$. The spacing for training is resized to $1 \times 1 \times 1$ for memory reasons. Intensities are normalised between 0 and 1 and data augmentation is performed before the training on the patches (flips, rotations, intensity). During inference, we post-process the output of all compared methods to identify the largest predicted connected component and fill any holes.\\

\textbf{Evaluation Metrics}
The first objective of this project is to recover the volume of each muscle. To this end, we mesure the volumetric error of each muscle, in $cm^3$ and percentage as:

\begin{equation}\label{errorCm3}
\begin{split}
    \textrm{Vol}_{err_{cm^3}}= |V_{GT}-V_{pred}| 
\end{split}
\qquad
\begin{split}
    \textrm{Vol}_{err_{\%}}= 100\times \frac{|V_{GT}-V_{pred}| }{V_{GT}} ,
\end{split}   
\end{equation}

where $V_{GT}$ and $V_{pred}$ correspond to the ground-truth and predicted volumes of a given muscle respectively. Note that $\textrm{Vol}_{err_{\%}}$ takes into account the size of the muscle from which we are extracting the volume, which $\textrm{Vol}_{err_{cm^3}}$ does not. 
We also rely on the Dice Score (DSC) and the 95$\%$ Hausdorff Distance (HD95) to evaluate the performance of the model.\\

\textbf{Implementation details}
The implementation relies on MONAI, a PyTorch-based open-source framework\footnote{https://monai.io/}.
Training was done on an NVIDIA GeForce RTX 3090 Ti (24 GB) graphic card. 
Training included two phases. During the first phase, the model was trained without NonAdjLoss for $6667$ epochs. Then, the model was fine-tuned with the regularisation loss for $5 000$ epochs. The regularization weight was set to $\lambda = 0.3$. 
Each model (pretrained and fine-tuned) was trained with a batch size of 1, using the AdamW optimizer and an initial learning rate of $0.0001$. Full training took $48$ hours ($11667$ epochs). The UNETR architecture was configured with $12$ transformer blocks ($T=12$) and has an embedding size of $K=768$ \cite{unetr}. To match the size of the data, we set the patch size to $128\times128\times128$.

\subsection{Quantitative results}
\begin{figure}[b!]
\includegraphics[width=0.8\textwidth]{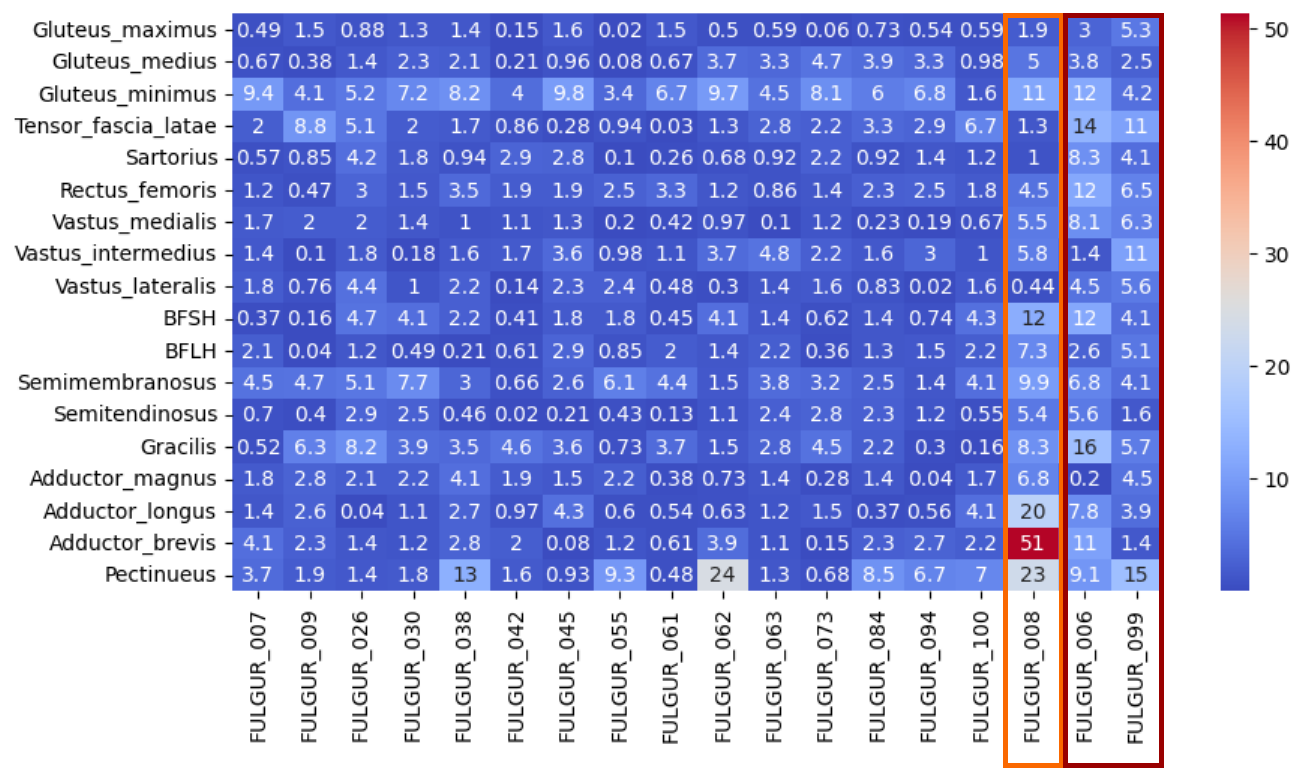}
\centering
\caption{Heatmap of the volumetric error ($\%$) of each muscle of each subject of the
database. The orange insert shows the result for the validation data and the red one
for the test set} \label{heatmap}
\end{figure}

\begin{figure}[t!]
\includegraphics[width=\textwidth]{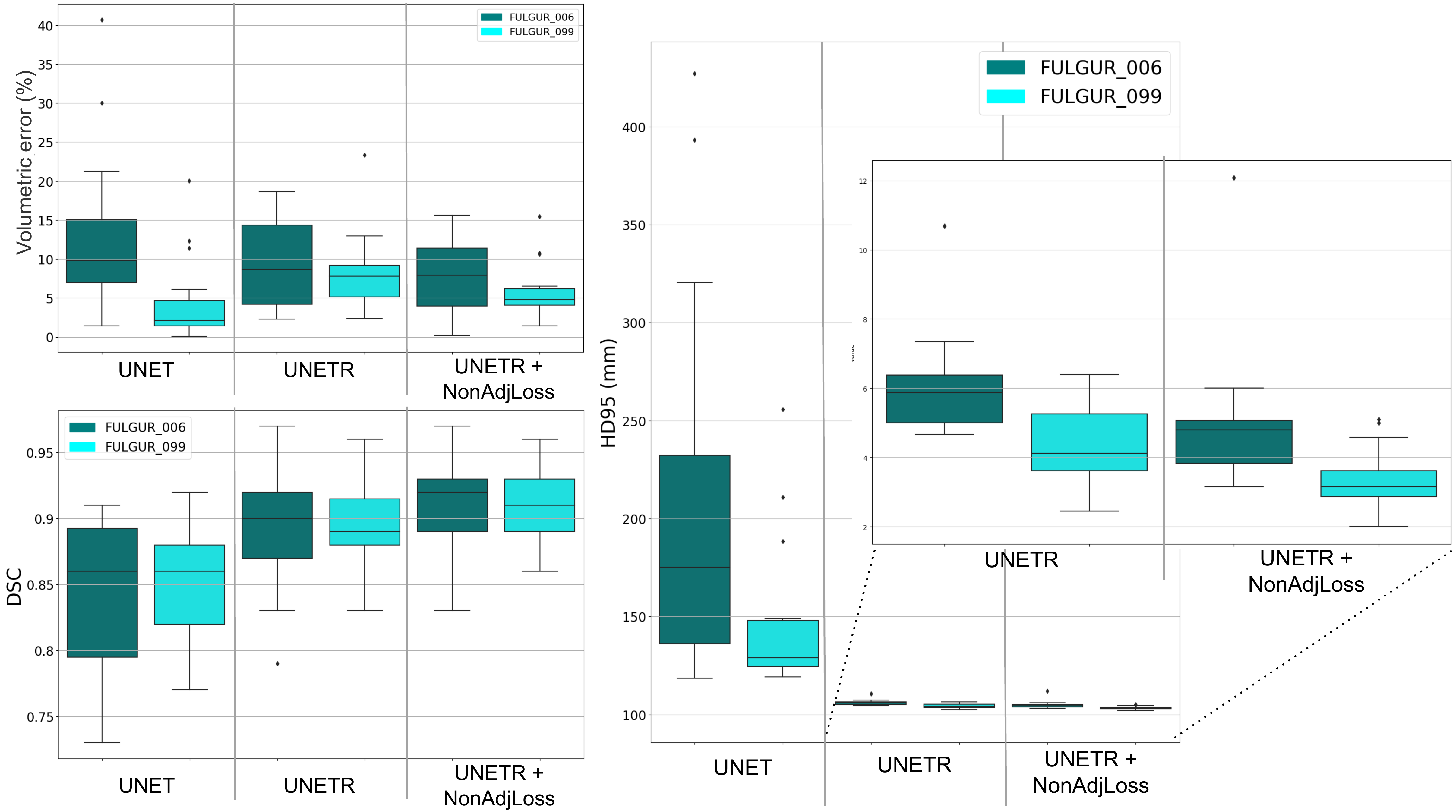}
\caption{Boxplot of the test results for UNET, UNETR and UNETR + NonAdjLoss. } \label{boxplots}
\end{figure}

Regarding the volumetric error ($\%$), we report the results in Fig.~\ref{heatmap}.
Most of the muscles have an error under $5 \%$ for the training set. 
As expected, the values for the test set are higher but in average bellow 10$\%$. Higher errors for the Pectineus can be explained by its small size and for the Gluteus minimus by its more challenging boundaries. 
In addition, we compared our method against a U-Net architecture. We also  investigated the impact of the regularization cost function on the learning process. To visualize the results, we present boxplots of the volumetric error, Dice coefficient, and Hausdorff distance 95 in Fig.~\ref{boxplots}, on the test dataset. 

The methods based on UNETR show an overall decrease in the number of outliers across metrics, indicating improved performance in terms of reducing extreme errors. Moreover, both UNETR and the fine-tuned method with NonAdjLoss reduce either the mean volumetric error or its variance. 
The transition to UNETR also results in a reduction of the 95HD error. For example, in the case of F006, the average Hausdorff distance decreased from approximately 175mm to around 6mm with UNETR. Similarly, for F099, the average Hausdorff distance was reduced from 125mm to an average of 4mm with UNETR. 
Moreover, the NonAdjLoss regularization further reduces the average Hausdorff distance, resulting in an average of 5mm for F006 and 3mm for F099. 
These results confirm that the inclusion of anatomical constraints enabled the learning process to generate predictions closer to the ground truth. The NonAdjLoss regularization has effectively guided the model to capture the anatomical characteristics and spatial relationships of the muscles, leading to improved segmentations.
Finally, there is an increase in Dice (DSC) with the methods incorporating transformers. We observe an average DSC of around 0.86 for U-Net, while U-Net with transformers (UNETR) achieves an average DSC of 0.9. Furthermore, when combining UNETR with NonAdjLoss, the average DSC further improves to 0.92. These results highlight the enhanced segmentation performance.

\subsection{Qualitative results}

\begin{figure}[t!]
\includegraphics[width=0.8\textwidth]{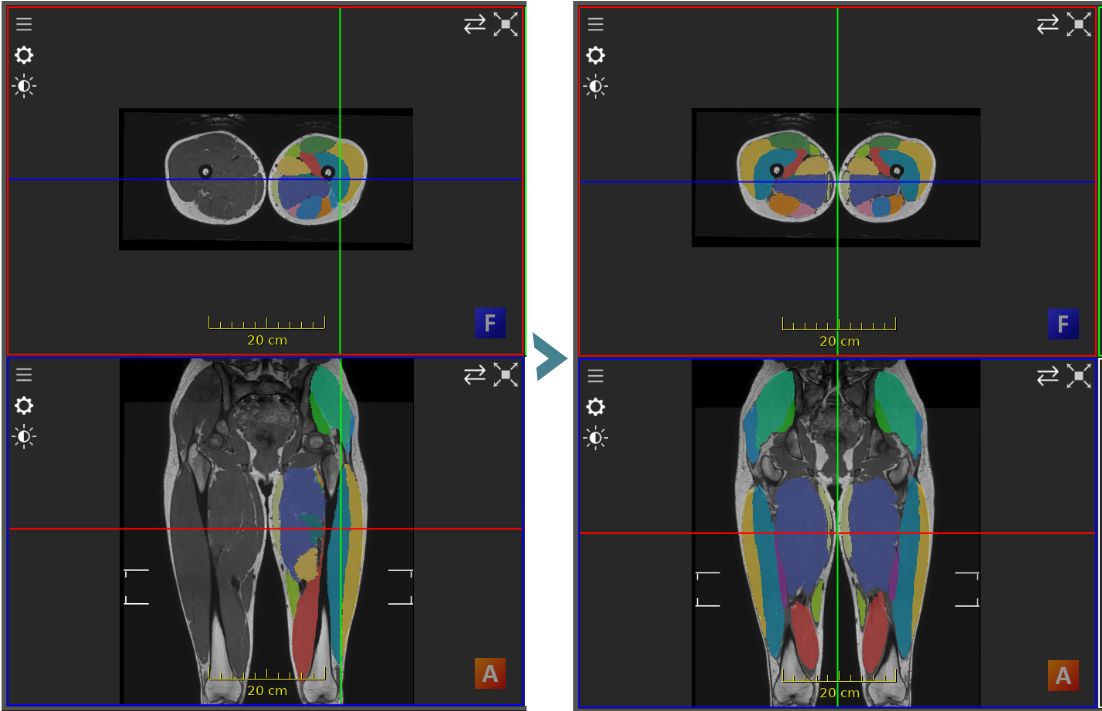}
\centering
\caption{Comparison of the Ground Truth (\textbf{left}) given as input to the model and the prediction of our trained model when we give an MRI with both legs (\textbf{right}).} \label{bothlegs}
\end{figure}

Regarding the qualitative results, one notable observation is that when providing the MRI scans of both legs, the model is capable of segmenting both legs successfully, even if it was only trained on one of them as shown in Fig.~\ref{bothlegs}. This can be attributed to the inherent symmetry found in human anatomy, the utilization of data augmentations during training
and the sequential inference process. The model has learned to generalize well to the other leg, leveraging the common features and structures 
These results demonstrate the adaptability of the model to handle variations such as symmetries and multiple instances. 

\begin{figure}[t!]
\includegraphics[width=\textwidth]{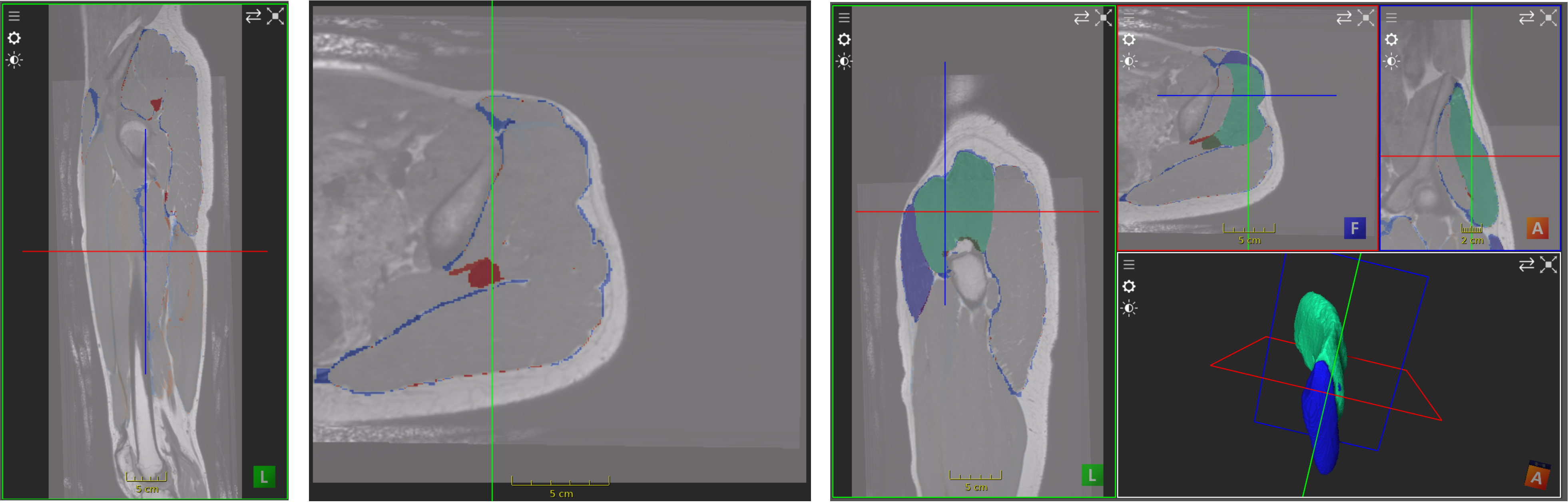}
\centering
\caption{Comparison of the predicted labelmap on a test subject and the GT labelmap, where blue color is the missing volume from the prediction and red color is the volume that is added comparatively to the GT (\textbf{left}). Prediction of the TFL(electric blue) and gluteus medius (turquoise green) that influence each other, supperposed with the comparison from the left part (\textbf{right}).
} \label{comparison}
\end{figure}

When we examine the predictions, we observe plausible and overall good quality labelmaps. The remaining errors, resemble human mistakes made during the  manual segmentation, which unfortunately can still be found in the ground truth of this dataset. 
Such errors include, voxels belonging to other labels be present within a muscle, and mixing the boundaries of muscles belonging to a group (such as adductor groups). Fig.~\ref{comparison}-right shows an example of the TFL that influences the gluteus maximus in the prediction, which further explains the quantitative results for that muscle. 
Additionally, we observe less anatomically accurate ground-truth label shapes in small ambiguous regions.
Therefore, the presence of some errors in both the manual annotations and the model predictions is expected. 

We can also observe that most errors occur at the boundaries of the segmentations as shown in Fig.~\ref{comparison}-left. This is particularly noticeable in the case of athletes since they have a significant muscle development, and the presence of adipose tissue between their muscles is reduced. However, suboptimal predictions can have a direct impact on adjacent predictions. For example, if a muscle is segmented slightly outside its boundaries, its neighboring muscle will have a reduced segmentation, resulting in a predicted decrease in muscle volume. This phenomenon occurs particularly in muscle groups such as the adductors (longus, magnus and brevis), where the boundaries are difficult to discern even to the naked eye and remain a challenge even to human experts.

\section{Conclusion}

We have proposed a method that leverages long-range shape dependencies and prior anatomical information to segment muscles of elite athletes. 
Our experimental validation demonstrates that by incorporating anatomical priors as constraints into the segmentation process, our method achieves improved accuracy and captures the nuances of muscle boundaries more effectively.
By accurately segmenting muscles, our method provides a valuable tool for quantitative analysis, allowing for a
more comprehensive assessments of muscle morphology. 
This information can be valuable in identifying potential asymmetries or variations, guiding personalized training programs, injury prevention strategies, and performance optimization in sports and athletic settings. 
Moreover, a prediction is significantly faster that  the manual segmentation method initially applied, which took approximately 30 to 40 hours per subject. 
Since, the significant amount of time required for manual segmentation and the challenging nature of the task have limited the size of the database, we plan to evaluate the revision time when starting from our method's predictions to confirm the acceleration of the labeling process for new subjects.\\

To further advance and explore potential improvements, several directions can be considered. 
The exploration of methods based on unlabelled data could be pursued. 
Another possibility is to investigate the application of newer architecture designs that combine CNNs and transformers, such as Swin Transformers \cite{swin_transformers}. 
A third avenue for improvement is to explore the reduction of the number of transformer blocks in the network \cite{Hasany_et_al}, to gain insights on the optimal balance between model complexity and segmentation accuracy. 
Finally, we plan to study the learned positional encodings and attention maps to better understand where the model focuses during the segmentation process. Analyzing attention maps can provide valuable insights into the features and regions that contribute most significantly to accurate muscle segmentation and to identify correlations between muscle groups.
This understanding can guide future refinement of the model architecture and its performance optimization. 
Finally, with some adaptions our method could be used for the morphological
study of muscles from patient with muskuloskeletal diseases.\\

\textbf{Acknowledgments. }

We would like to acknowledge the manual segmentation operators, Iwen Dirouon and Eva Filleur, for their invaluable contribution. We would also like to thank Caroline Giroux for her support with the database. In conclusion, our gratitude extends to Guillaume Pelluet for his collaboration in developing the adjacency matrix code.  

% ---- Bibliography ----
\bibliographystyle{splncs04}
\bibliography{biblio-macro,bibliography}

\end{document}